\documentclass{article} % For LaTeX2e
\usepackage[final]{nips}

\usepackage{hyperref}

\usepackage{times}
\usepackage[utf8]{inputenc} % allow utf-8 input
\usepackage[T1]{fontenc}    % use 8-bit T1 fonts
\usepackage{hyperref}       % hyperlinks
\usepackage{url}            % simple URL typesetting
\usepackage{booktabs}       % professional-quality tables
\usepackage{amsmath, amsfonts}       % blackboard math symbols
\usepackage{nicefrac}       % compact symbols for 1/2, etc.
\usepackage{microtype}      % microtypography
\usepackage{graphicx}
\usepackage{wrapfig}
\usepackage{color}
\usepackage{enumitem}
\usepackage[font={small}]{caption}
\DeclareCaptionType{noticebox}
\usepackage{subcaption}
\usepackage{epstopdf}
\usepackage{multirow}
\title{An empirical study on evaluation metrics of generative adversarial networks}

\author{Qiantong Xu, Gao Huang, Yang Yuan, Chuan Guo \\
Cornell University\\
\texttt{\{qx57, gh349, yy528, cg563\}@cornell.edu} 
\And
Yu Sun\\
UC Berkeley\\
\texttt{yusun@berkeley.edu} \\
\And
Felix Wu, Kilian Q. Weinberger\\
Cornell University\\
\texttt{\{fw245, kqw4\}@cornell.edu} \\
}

\renewcommand{\paragraph}[1]{

\textbf{#1}}

\newcommand{\bx}{\ensuremath{\mathbf{x}}}

\newcommand{\bz}{\ensuremath{\mathbf{z}}}
\newcommand{\bmu}{\ensuremath{\mathbf{\mu}}}
\newcommand{\bC}{\ensuremath{\mathbf{C}}}

\DeclareMathOperator{\IS}{IS}
\DeclareMathOperator{\MS}{MS}
\DeclareMathOperator{\MMD}{MMD}
\DeclareMathOperator{\WD}{WD}
\DeclareMathOperator{\FID}{FID}
\newcommand{\eat}[1]{}

\begin{document}

\maketitle
\begin{abstract}
%!TEX root=main.tex
%Despite the widespread interest in generative adversarial networks (GANs), 
Evaluating generative adversarial networks~(GANs) is inherently challenging. 
%Yet, to this date,  have studied the metrics that quantitatively evaluate GANs' performance.
In this paper, we revisit several representative sample-based evaluation metrics for GANs, and address the problem of \emph{how to evaluate the evaluation metrics}.
We start with a few necessary conditions for metrics to produce meaningful scores, such as distinguishing real from generated samples, identifying mode dropping and mode collapsing, and detecting overfitting.
With a series of carefully designed experiments,  we comprehensively investigate existing sample-based metrics and identify their strengths and limitations in practical settings.
Based on these results, we observe that kernel Maximum Mean Discrepancy (MMD) and the 1-Nearest-Neighbor (1-NN) two-sample test seem to satisfy most of the desirable properties, \emph{provided that the distances between samples are computed in a suitable feature space}.
Our experiments also unveil interesting properties about the behavior of several popular GAN models, such as whether they are memorizing training samples, and how far they are from learning the target distribution. 
%these state-of-the-art GANs are from perfect.

\end{abstract}

%!TEX root=main.tex
\vspace{-4 ex}
\section{Introduction}
\vspace{-1 ex}
Generative adversarial networks (GANs) \citep{goodfellow2014generative} have been studied extensively in recent years. Besides producing surprisingly plausible images~\citep{radford2015unsupervised,larsen2015autoencoding,karras2017progressive,arjovsky2017wasserstein,gulrajani2017improved}, they have also been innovatively applied in, for example, semi-supervised learning \citep{odena2016semi,makhzani2015adversarial}, image-to-image translation \citep{isola2016image,zhu2017unpaired}, and simulated image refinement \citep{shrivastava2016learning}.
However, despite the availability of a plethora of GAN models \citep{arjovsky2017wasserstein,qi2017lsgan,zhao2016ebgan}, their evaluation is still predominantly qualitative, very often resorting to manual inspection of the visual fidelity of generated images. Such evaluation is time-consuming, subjective, and possibly misleading. Given the inherent limitations of \emph{qualitative} evaluations, proper \emph{quantitative} metrics are crucial for the development of GANs to  guide the design of better models.

Possibly the most popular metric is the Inception Score \citep{salimans2016improved}, which measures the quality and diversity of the generated images using an external model, the Google Inception network \citep{szegedy2014inception}, trained on the large scale ImageNet dataset \citep{deng2009imagenet}. Some other metrics are less widely used but still very valuable.
\citet{wu2016quantitative} proposed a sampling method to estimate the log-likelihood of generative models, by assuming a Gaussian observation model with a fixed variance. \citet{bounliphone2015test} propose to use maximum mean discrepancies (MMDs) for model selection in generative models.
\citet{lopez2016revisiting} apply the classifier two-sample test, a well-studied tool in statistics, to assess the difference between the generated and target distribution.

Although these evaluation metrics are shown to be effective on various tasks, it is unclear in which scenarios their scores are meaningful, and in which other scenarios prone to misinterpretations.
Given that evaluating GANs is already challenging it can only be more difficult to evaluate the evaluation metrics themselves.
Most existing works attempt to justify their proposed metrics by showing a strong correlation with human evaluation~\citep{salimans2016improved,lopez2016revisiting}. However, human evaluation tends to be biased towards the visual quality of generated samples and neglect the overall distributional characteristics, which are important for unsupervised learning.

In this paper we comprehensively examine the existing literature on sample-based quantitative evaluation of GANs. We address the challenge of evaluating the metrics themselves by carefully designing a series of experiments through which we hope to answer the following questions:
1) What are reasonable characterizations of the behavior of existing sample-based metrics for GANs?
2) What are the strengths and limitations of these metrics, and which metrics should be preferred accordingly?
%4.) How are the metrics helpful in understanding and improving GANs?
Our empirical observation suggests that MMD and 1-NN two-sample test are best suited as evaluation metrics on the basis of satisfying useful properties such as discriminating real versus fake images, sensitivity to mode dropping and collapse, and computational efficiency.

Ultimately, we hope that this paper will establish good principles on choosing, interpreting, and designing evaluation metrics for GANs in practical settings. We will also release the source code for all experiments and metrics examined (\url{https://github.com/xuqiantong/GAN-Metrics}), providing the community with off-the-shelf tools to debug and improve their GAN algorithms.

%!TEX root=main.tex
\vspace{-1 ex}
\section{Background}
\label{related}
\vspace{-1 ex}

We briefly review the original GAN framework proposed by \citet{goodfellow2014generative}.  Description of the GAN variants used in our experiments is deferred to the Appendix \ref{sec:gan-variants}.

\vspace{-1 ex}
\subsection{Generative adversarial networks}
\vspace{-1 ex}
Let $\mathcal{X} = \mathbb{R}^{d\times d}$ be the space of natural images. Given i.i.d. samples $S_r = \{\bx_1^r,\ldots,\bx_n^r\}$ drawn from a real distribution $\mathbb{P}_r$ over $\mathcal{X}$, we would like to learn a parameterized distribution $\mathbb{P}_g$ that approximates the distribution $\mathbb{P}_r$.

A generative adversarial network has two components, the discriminator $D : \mathcal{X} \rightarrow [0,1)$ and the generator $G : \mathcal{Z} \rightarrow \mathcal{X}$, where $\mathcal{Z}$ is some latent space. Given a distribution $\mathbb{P}_z$ over $\mathcal{Z}$ (usually an isotropic Gaussian),
the distribution $\mathbb{P}_g$ is defined as $G(\mathbb{P}_z)$. % WTF CHUAN, YOU CAN'T DO THIS...
Optimization is performed with respect to a joint loss for $D$ and $G$
\begin{equation}
\vspace{-1ex}
\min_G \max_D  \ell \!=\! \small\text{$\mathbb{E}_{\bx \sim \mathbb{P}_r} \log \left[D(\bx)\right]
+ \mathbb{E}_{\bz \sim \mathbb{P}_z} \left[\log(1 - D(G(\bz)))\right]$}. \nonumber
\vspace{-1ex}
\end{equation}
%\vspace{-1ex}

Intuitively, the discriminator $D$ outputs a probability for every $\bx \in \mathcal{X}$ that corresponds to its likelihood of being drawn from $\mathbb{P}_r$, and the loss function encourages the generator $G$ to produce samples that maximize this probability. Practically, the loss is approximated with finite samples from $\mathbb{P}_r$ and $\mathbb{P}_g$, and optimized with alternating steps for $D$ and $G$ using gradient descent.

To evaluate the generator, we would like to design a metric $\rho$ that measures the ``dissimilarity" between $\mathbb{P}_g$ to $\mathbb{P}_r$.\footnote{Note that $\rho$ does not need satisfy symmetry or triangle inequality, so it is not, mathematically speaking, a distance metric between
$\mathbb{P}_g$ and $\mathbb{P}_r$. We still call it a metric throughout this paper for simplicity.}
In theory, with both distributions known, common choices of $\rho$ include the Kullback-Leibler divergence (KLD), Jensen-Shannon divergence (JSD) and total variation.
However, in practical scenarios, $\mathbb{P}_r$ is unknown and only the finite samples in $S_r$ are observed.
Furthermore, it is almost always intractable to compute the exact density of $\mathbb{P}_g$, but much easier to sample $S_g = \{\bx_1^g,\ldots,\bx_m^g\} \sim \mathbb{P}_g^m$ (especially so for GANs).
Given these limitations, we focus on \emph{empirical} measures $\hat{\rho} : \mathcal{X}^n \times \mathcal{X}^m \rightarrow \mathbb{R}$ of ``dissimilarity" between samples from two distributions.

\begin{figure*}[t]
	\centering
	\includegraphics[width=0.8\textwidth]{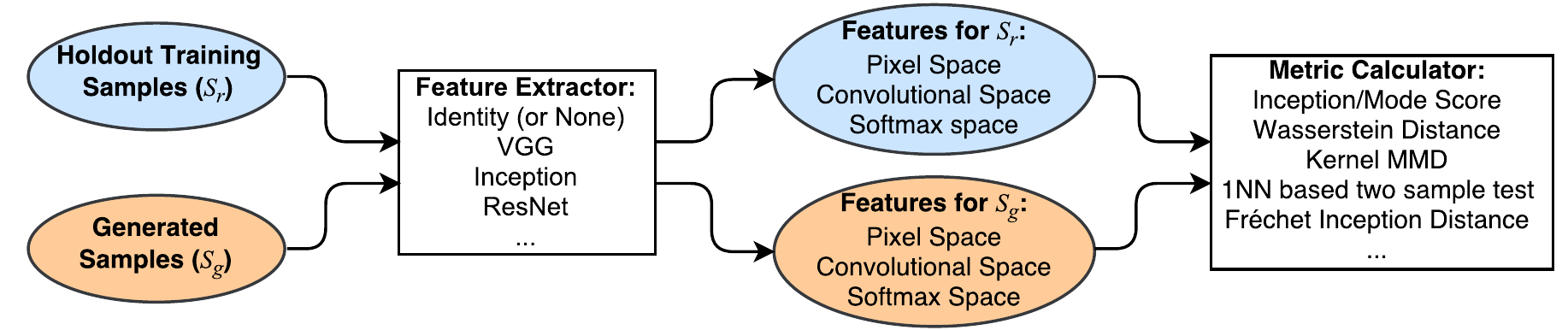}
	\vspace{-1 ex}
	\caption{Typical sample based GAN evaluation methods. 	}
	\label{fig:work-flow}
	\vspace{-2 ex}
\end{figure*}

\vspace{-1 ex}
\subsection{Sample based metrics}
\label{subsec:metrics}
\vspace{-1 ex}

We mainly focus on sample based evaluation metrics that follow a common setup illustrated in \autoref{fig:work-flow}. The \emph{metric calculator} is the key element, for which we briefly introduce five representative methods:
Inception Score~\citep{salimans2016improved},
Mode Score~\citep{che2016mode} ,
Kernel MMD~\citep{gretton2007kernel},
Wasserstein distance, Fr\'echet Inception Distance (FID)~\citep{heusel2017gans}, and 1-nearest neighbor (1-NN)-based two sample test~\citep{lopez2016revisiting}.
All of them are model agnostic and require only finite samples from the generator.

\paragraph{The Inception Score} is arguably the most widely adopted metric in the literature. It uses a image classification model $\mathcal{M}$, the Inception network \citep{szegedy2016rethinking}, pre-trained on the ImageNet~\citep{deng2009imagenet} dataset, to compute
\begin{align}
\label{eq:incep}
\vspace{-1ex}
\IS(\mathbb{P}_g) = e^{\mathbb{E}_{\bx \sim \mathbb{P}_g} \left[{KL}(p_\mathcal{M}(y|\bx)||p_\mathcal{M}(y))\right]},\nonumber
\vspace{-1ex}
\end{align}
where $p_\mathcal{M}(y|\bx)$ denotes the label distribution of $\mathbf{x}$ as predicted by $\mathcal{M}$,
and {$p_\mathcal{M}(y) = \int_\bx p_\mathcal{M}(y|\bx)~d\mathbb{P}_g$},
i.e. the marginal of $p_\mathcal{M}(y|\bx)$ over the probability measure $\mathbb{P}_g$.
The expectation and the integral in $p_\mathcal{M}(y|\bx)$ can be approximated with i.i.d. samples from $\mathbb{P}_g$. A higher $\IS$ has $p_\mathcal{M}(y|\bx)$ close to a point mass, which happens when the Inception network is very confident that the image belongs to a particular ImageNet category, and has $p_\mathcal{M}(y)$ close to uniform, i.e. all categories are equally represented. This suggests that the generative model has both high quality and diversity.
\citet{salimans2016improved} show that the Inception Score has a reasonable correlation with human judgment of image quality.
We would like to highlight two specific properties: 1) the distributions on both sides of the KL are dependent on $\mathcal{M}$,  and 2) the distribution of the real data $\mathbb{P}_r$, or even samples thereof,  are not used anywhere.

\paragraph{The Mode Score\footnote{We use a modified version here, as the original one reduces to the Inception Score.}} is an improved version of the Inception Score. Formally, it is given by
\begin{equation}
\label{eq:mode-score}
\vspace{-1ex}
\MS(\mathbb{P}_g) \!=\! e^{\mathbb{E}_{\bx \sim \mathbb{P}_g} [{KL}(p_\mathcal{M}(y|\bx)||p_\mathcal{M}(y))] \!-\! KL( p_\mathcal{M}(y)||p_\mathcal{M}(y^*))},\nonumber
%\vspace{-1ex}
\end{equation}
where  $p_\mathcal{M}(y^*) = \int_\bx p_\mathcal{M}(y|\bx)~d\mathbb{P}_r$ is the marginal label distribution for the samples from the real data distribution.
Unlike the Inception Score, it is able to measure the dissimilarity between the real distribution $\mathbb P_r$
and generated distribution $\mathbb P_g$ through the term $KL( p_\mathcal{M}(y)||p_\mathcal{M}(y^*))$.

\paragraph{Kernel MMD} (Maximum Mean Discrepancy) is defined as
\begin{equation}
\label{eq:mmd}
\vspace{-1ex}
\MMD^2(\!\mathbb{P}_r,\mathbb{P}_g\!)\!=\!
\mathbb E_{\substack{\bx_r, \bx_r' \sim \mathbb{P}_r,\\
	 \bx_g, \bx_g' \sim \mathbb{P}_g}}
\tiny\text{$ \bigg[k(\bx_r,\bx_r')\!-\!2k(\bx_r,\bx_g)\!+\!k(\bx_g,\bx_g')\bigg]$},\nonumber
%\vspace{-1ex}
\end{equation}
measures the dissimilarity between $\mathbb{P}_r$ and $\mathbb{P}_g$ for some fixed kernel function $k$.
Given two sets of samples from $\mathbb{P}_r$ and $\mathbb{P}_g$, the empirical MMD between the two distributions can be computed with finite sample approximation of the expectation. A lower MMD means that $\mathbb{P}_g$ is closer to $\mathbb{P}_r$. The Parzen window estimate \citep{gretton2007kernel} can be viewed as a specialization of Kernel MMD.

\paragraph{The Wasserstein distance} between $\mathbb{P}_r$ and $\mathbb{P}_g$ is defined as
\begin{equation}
\label{eq:emd}
\WD(\mathbb{P}_r, \mathbb{P}_g) = \inf_{\gamma\in \Gamma(\mathbb{P}_r, \mathbb{P}_g)}\mathbb E_{(\bx^r,\bx^g)\sim\gamma} \left[d(\bx^r,\bx^g)\right], \nonumber
\end{equation}
where $\Gamma(\mathbb{P}_r, \mathbb{P}_g)$ denotes the set of all joint distributions (i.e. probabilistic couplings) whose marginals are respectively $\mathbb{P}_r$ and $\mathbb{P}_g$, and $d(\bx^r, \bx^g)$ denotes the base distance between the two samples. For discrete distributions with densities $p_r$ and $p_g$, the Wasserstein distance is often referred to as the Earth Mover's Distance (EMD), and corresponds to the solution to the optimal transport problem
\begin{align}
\WD(p_r, p_g) \!=\! \min\nolimits_{w \in \mathbb{R}^{n \times m}} \sum\nolimits_{i=1}^n \sum\nolimits_{j=1}^m w_{ij} d(\bx_i^r, \bx_j^g)\quad \nonumber \\
\text{s.t.}~~\sum\nolimits_{j=1}^m w_{i,j} \!=\! p_r(\bx_i^r)~~\forall i,
\sum\nolimits_{i=1}^n w_{i,j} \!=\! p_g(\bx_j^g)~~\forall j. \nonumber
\end{align}
This is the finite sample approximation of $\WD(\mathbb{P}_r, \mathbb{P}_g)$ used in practice. Similar to MMD, the Wasserstein distance is lower when two distributions are more similar.

\paragraph{The Fr\'echet Inception Distance (FID)} was recently introduced by \citet{heusel2017gans} to evaluate GANs. For a suitable feature function $\phi$ (by default, the Inception network's convolutional feature), FID models $\phi(\mathbb{P}_r)$ and $\phi(\mathbb{P}_g)$ as Gaussian random variables with empirical means $\mu_r, \mu_g$ and empirical covariance $\bC_r, \bC_g$, and computes
\begin{equation}
\FID(\mathbb{P}_r, \mathbb{P}_g) = \|\bmu_r-\bmu_g\|+\text{Tr}(\bC_r+\bC_g-2(\bC_r\bC_g)^{1/2}),\nonumber
\end{equation}
which is the Fr\'echet distance (or equivalently, the Wasserstein-2 distance) between the two Gaussian distributions \citep{heusel2017gans}.

\paragraph{The 1-Nearest Neighbor classifier} is used in two-sample tests to assess whether two distributions are identical.
Given two sets of samples $S_r \sim \mathbb{P}_r^n$ and $S_g \sim \mathbb{P}_g^m$, with $|S_r|=|S_g|$, one can compute the leave-one-out (LOO) accuracy of a 1-NN classifier trained on $S_r$ and $S_g$ with positive labels for $S_r$ and negative labels for $S_g$. Different from the most common use of accuracy, here the 1-NN classifier should yield a $\sim \!50\%$ LOO accuracy when $|S_r|=|S_g|$ is large. This is achieved when the two distributions match. The LOO accuracy can be lower than $50\%$, which happens when the GAN overfits $\mathbb{P}_g$ to $S_r$. In the (hypothetical) extreme case, if the GAN were to memorize every sample in $S_r$ and re-generate it exactly, i.e. $S_g=S_r$, the accuracy would be $0\%$, as every sample from $S_r$ would have it nearest neighbor from $S_g$ with zero distance. The 1-NN classifier belongs to the two-sample test family, for which any binary classifier can be adopted in principle. We will only consider the 1-NN classifier because it requires no special training and little hyperparameter tuning.

\citet{lopez2016revisiting} considered the 1-NN accuracy primarily as a statistic for two-sample testing. In fact, it is more informative to analyze it for the two classes separately. For example, a typical outcome of GANs is that for both real \emph{and} generated images, the majority of their nearest neighbors are generated images due to mode collapse. In this case, the LOO 1-NN accuracy of the real images would be relatively low (desired): the mode(s) of the real distribution are usually well captured by the generative model, so a majority of real samples from $S_r$ are surrounded by generated samples from $S_g$, leading to low LOO accuracy; whereas the LOO accuracy of the generated images is high (not desired): generative samples tend to collapse to a few mode centers, thus they are surrounded by samples from the same class, leading to high LOO accuracy.
For the rest of the paper, we distinguish these two cases as 1-NN accuracy (real) and 1-NN accuracy (fake).

%Intuitively, such a setting happens in high dimensional spaces when the real images are spread out sparsely (more diverse) while the generated images are clustered around a few centers (mode collapsing).

%Improving upon the previous work, we find it more informative to break down the overall LOO accuracy, and look at that for real and generated samples separately.
%When the accuracy for real samples is high, it says that the generative samples are far from majority of the real samples.
%When the accuracy for generated samples is high, it says that they tend to cluster together, i.e. generated samples are mostly surrounded by generated ones in the distribution defined by $\mathcal{P}_g$. This is normally a symptom of model dropping or mode collapsing.
%{\color{red}We need to work a little harder on the interpretations - I still don't think they are intuitive enough. This break down is one of our novelties so we should put more insight into it.}

\vspace{-1 ex}
\subsection{Other metrics}
\vspace{-1 ex}
All of the metrics above are, what we refer to as ``\emph{model agnostic}": they use the generator as a black box to sample the generated images $S_g$.
Model agnostic metrics should not require a density estimation from the model. We choose to only experiment with model agnostic metrics, which allow us to support as many generative models as possible for evaluation without modification to their structure.
We will briefly mention some other evaluation metrics not included in our experiments.

Kernel density estimation (KDE, or Parzen window estimation) is a well-studied method for estimating the density function of a distribution from samples. For a probability kernel $K$ (most often an isotropic Gaussian) and i.i.d samples $\bx_1,\ldots,\bx_n$, we can define the density function at $\bx$ as
${p}(\bx) \approx \frac{1}{z} \sum_{i=1}^n K(\bx - \bx_i)$, where $z$ is a normalizing constant.
This allows the use of classical metrics such as KLD and JSD. However, despite the widespread adoption of this technique to various applications, its suitability to estimating the density of $\mathbb{P}_r$ or $\mathbb{P}_g$ for GANs has been questioned by \citet{theis2015note} since the probability kernel depends on the Euclidean distance between images.
%Moreover, \citet{gretton2012kernel} has shown that the $L_2$ distance between the density estimates of $\mathbb{P}_r$ and $\mathbb{P}_g$ using samples $S_r$ and $S_g$ is equivalent to a special case of MMD with the inner product kernel.

More recently, \citet{wu2016quantitative} applied annealed importance sampling (AIS) to estimate the marginal distribution $p(\bx)$ of a generative model. This method is most natural for models that define a conditional distribution $p(\bx | \bz)$ where $\bz$ is the latent code, which is not satisfied by most GAN models. Nevertheless, AIS has been applied to GAN evaluation by assuming a Gaussian observation model. We exclude this method from our experiments as it needs the access to the generative model to compute the likelihood, instead of only depending on a finite sample set $S_g$.

\vspace{-1 ex}
\section{Experiments with GAN evaluation metrics}
\label{experiments}
\vspace{-1 ex}

\subsection{Feature space}
\vspace{-1 ex}

%\vspace{-2 ex}
%\begin{wrapfigure}{r}{0.41\textwidth}
%      \vspace{-4 ex}
%      \centering
%      \includegraphics[width=0.2\textwidth]{figures/celeba_10nn_shift4_pixel_7x7.png}
%      \includegraphics[width=0.2\textwidth]{figures/celeba_10nn_shift4_conv_7x7.png}\\
%            %\includegraphics[width=0.5\textwidth]{figures/visual_earlyexit_cifar10-2.pdf}
%      \caption{Shifted images (in green box) and their 10 nearest neighbors in the pixel space (\emph{left}) and in the ResNet feature space (\emph{right}).}
%      \vspace{-2 ex}
%      \label{fig:robust_shift}
%\end{wrapfigure}

All the metrics introduced in the previous section, except for the Inception Score and Mode Score, access the samples $\bx$ only through pair-wise distances. The Kernel MMD requires a fixed kernel function $k$, typically set to an isotopic Gaussian; the Wasserstein distance and 1-NN accuracy use the underlying distance metric $d$ directly; all of these methods are highly sensitive to the choice that distance.

It is well-established that pixel representations of images do not induce meaningful Euclidean distances~\citep{forsyth2011computer}.
Small translations, rotations, or changes in illumination can increase distances dramatically with little effect on the image content.
To quantity the similarity between distributions of images, it is therefore desirable to use distances invariant to such transformations.
The choice of distance function can be re-interpreted as a choice of representation, by defining the distance in a more general form as $d(\bx,\bx')=\|\phi(\bx)-\phi(\bx')\|_2$, where $\phi(\cdot)$ is some general mapping of the input into a semantically meaningful feature space. For Kernel MMD, this corresponds to computing the usual inner product in the feature space $\phi(\mathcal{X})$.

% The choice of this mapping $\phi$ controls the degree to which \emph{the induced distance is sensitive to identity-preserving transformations}.
% Since $\mathbb{P}_r$ is supported over the natural images, we expect $\mathbb{P}_r$ to have certain invariants, e.g. if $\bx, \bx' \in \mathcal{X}$ are the same up to a small translation or a local distortion (small diffeomorphism), we should have $p_r(\bx) \approx p_r(\bx')$ where $p_r$ is the density of $\mathbb{P}_r$. Thus if we apply such transformations to the samples in $S_r$ to produce $S_r'$, we expect that $\hat{\rho}(S_r,S_r') \approx 0$.

%We now consider the invariance properties of different choices of feature spaces. The most straightforward is the raw pixel space $\mathcal{X}$, corresponding to $\phi$ being the identity mapping. As expected, the distance induced by the pixel space is very sensitive. This makes distance-based metric problematic, as also addressed in \citep{theis2015note}. It partly explains why those well-known metrics not been adopted for evaluating GANs.

Inspired by the Inception/Mode Score, and recent works from~\cite{UpchurchGBPSW16,larsen2015autoencoding} which show that convolutional networks may linearize the image manifold, we propose to operate in the feature space of a pre-trained model on the ImageNet dataset.
For efficiency, we use a 34-layer ResNet as the feature extractor. Our experiments show that other models such as VGG or Inception give very similar results.

%\autoref{fig:robust_shift} shows a simple comparison of distances in the induced space and the original pixel representation.
%We randomly select images from CelebA (see dataset description in \autoref{experiments}), shift them by 4 pixels to the left and find their 10 nearest neighbors in the original dataset. Since the shifted images are semantically identical to the original, we expect a distance metric in a semantically meaningful feature space to include the original as the nearest neighbor.
%As expected, this is realized in the convolutional feature space (right plot), but completely unsuccessful in the pixel and softmax space. A similar observation holds for horizontal flips (see Figure 1 in the appendix).

To illustrate our point, we show failure examples of the pixel space distance for evaluating GANs in this section, and highlight that using a proper feature space is key to obtaining meaningful results when applying the distance-based metrics. The usage of a well-suited feature space enables us to draw more optimistic conclusions on GAN evaluation metrics than in \cite{theis2015note}.

\vspace{-1 ex}
\subsection{Setup}
\vspace{-1 ex}
For the rest of this major section, we introduce what in our opinion are necessary conditions for good metrics for GANs. After the introduction of each condition, we use it as a criterion to judge the effectiveness of the metrics presented in Section 2, through carefully designed empirical experiments.

The experiments are performed on two commonly used datasets for generative models, CelebA\footnote{CelebA is a large-scale high resolution face dataset with more than 200,000 centered celebrity images.} and LSUN bedrooms\footnote{LSUN consists of around one million images for each of 10 scene classes.
Following standard practice, we only take the \emph{bedroom} scene.}.
To remove the degree of freedom induced by feature representation, we use the Inception Score (IS) and Mode Score (MS) computed from the softmax probabilities of the same ResNet-34 model as the other metrics, instead of the Inception model.
We also compute the Inception Score over the real training data $S_r$ as an upper bound, which we denote as $\text{IS}_0$.
Moreover, to be consistent with other metrics where lower values correspond to better models, we report the \emph{relative inverse} Inception Score $RIS=(1\!-\!\text{IS}/\text{IS}_0)$ here, after computing $\text{IS}$ the Inception Score.
We similarly report the relative inverse Mode Score (RMS).
Although RIS and RMS operate in the softmax space, we always compare them together with other metrics in the convolutional space for simplicity. For all the plots in this paper, shaded areas denote the standard deviations, computed by running the same experiment $5$ times with different random seeds.

\begin{figure*}[t]
	\centering
	\vspace{-1 ex}
	\includegraphics[width=0.9\textwidth]{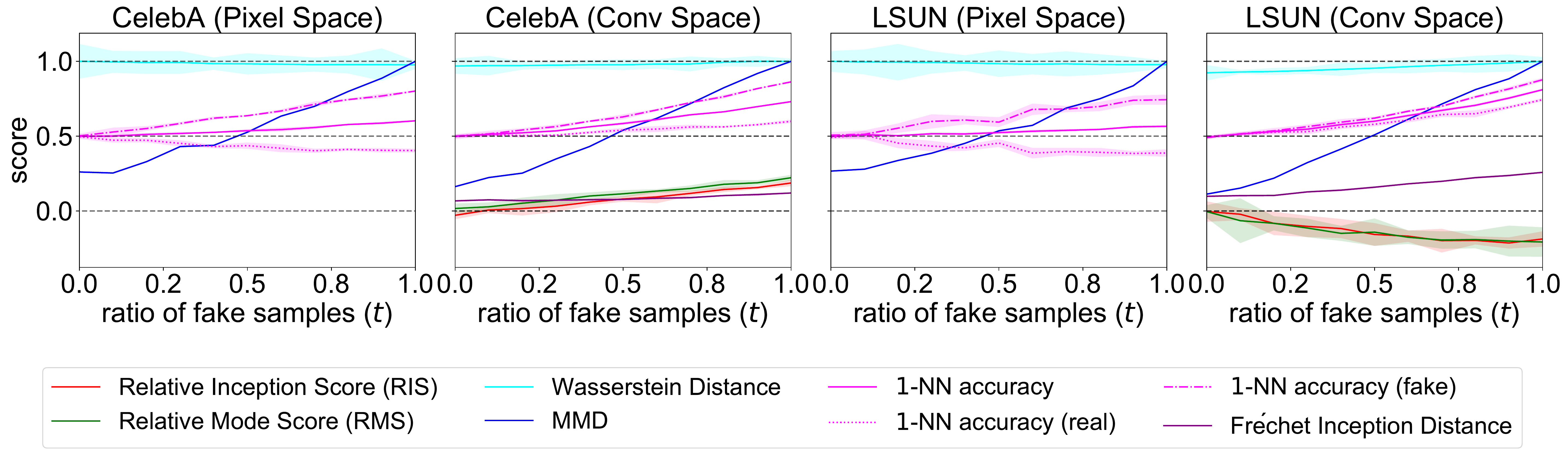}
	\vspace{-1 ex}
	\caption{Distinguishing a set of real images from a mixed set of real images and GAN generated images.
	For the metric to be discriminative, its score should increase as the fraction of generated samples in the mix increases.
	RIS and RMS fail as they decrease with the fraction of generated samples in $S_g$ on LSUN.
	Wasserstein and 1-NN accuracy (real) fail in pixel space as they do not increase. 	}
	\label{fig:mix-ratio}
	\vspace{-3 ex}
\end{figure*}

\vspace{-1 ex}
\subsection{Discriminability}
\vspace{-1 ex}
\paragraph{Mixing of generated images.}
Arguably, the most important property of a metric $\hat{\rho}$ for measuring GANs is the ability to distinguish generated images from real images. To test this property, we sample a set $S_r$ consisting of $n$ ($n=2000$ if not otherwise specified) real images uniformly from the training set,
and a set $S_g(t)$ of the same size $n$ consisting of a \emph{mix} of real samples and generated images from a DCGAN~\citep{radford2015unsupervised} trained on the same training set, where $t$ denotes the ratio of generated images.

The computed values of various metrics between $S_r$ and $S_g(t)$ for $t \in [0,1]$ are shown in \autoref{fig:mix-ratio}.
Since $\mathbb{P}_r$ should serve as a lower bound for any metric, we expect that any reasonable $\hat{\rho}$ should increase as the ratio of generated images increases. This is indeed satisfied for all the metrics \emph{except}: 1) RIS and RMS (red and green curves) on LSUN, %which ignores $S_r$ entirely;
which decrease as more fake samples are in the mix;
2) 1-NN accuracy of real samples (dotted magenta curve) computed in \emph{pixel space}, which also appears to be a decreasing function;
and 3) Wasserstein Distance (cyan curve), which almost remains unchanged when $t$ varies.

The reason that  RIS and RMS do not work well here is likely because they are not suitable for images beyond the ImageNet categories. Although other metrics operate in the convolutional feature space also depend on a network pretrained on ImageNet, the convolutional features are much more general than the specific softmax representation.
The failure of Wasserstein Distance is possibly due to an insufficient number of samples, which we will discuss in more detail when we analyze the sample efficiency of various metrics in a latter subsection. The last paragraph of Section~\ref{subsec:metrics} explains why the 1-NN accuracy for real samples (dotted magenta curve) is always lower than that for generated samples (dashed magenta curve). In the pixel space, more than half of the samples from $S_r$ have the nearest neighbor from $S_g(t)$, indicating that the DCGAN is able to represent the modes in the pixel space quite well.

We also conducted the same experiment using
1) random noise images and 2) images from an entirely different distribution (e.g. CIFAR-10), instead of DCGAN generated images to construct $S_g(t)$.
We call these injected samples as \emph{out-of-domain violations} since they are not in $\mathcal{X}$, the domain of the real images. These settings yield similar results as in \autoref{fig:mix-ratio}, thus we omit their plots.

%All the metrics except RIS and RMS are able to reflect the degree of out-of-domain violations, provided the distances are computed in the convolutional feature space, highlighting that choosing a proper feature space is crucial.

%In the Appendix, we repeat those experiments but use 1) random noise images and 2) images from an entirely different distribution (e.g. CIFAR-10), instead of DCGAN generated images. We call these injected samples as \emph{out-of-domain violations} since they are not in $\mathcal{X}$, the domain of the real images. All the metrics except RIS are able to reflect the degree of out-of-domain violations, provided the distances are computed in the convolutional feature space, highlighting that choosing a proper feature space is crucial.

\begin{figure*}[t]
	\centering
	\vspace{-1 ex}
    \includegraphics[width=0.9\textwidth]{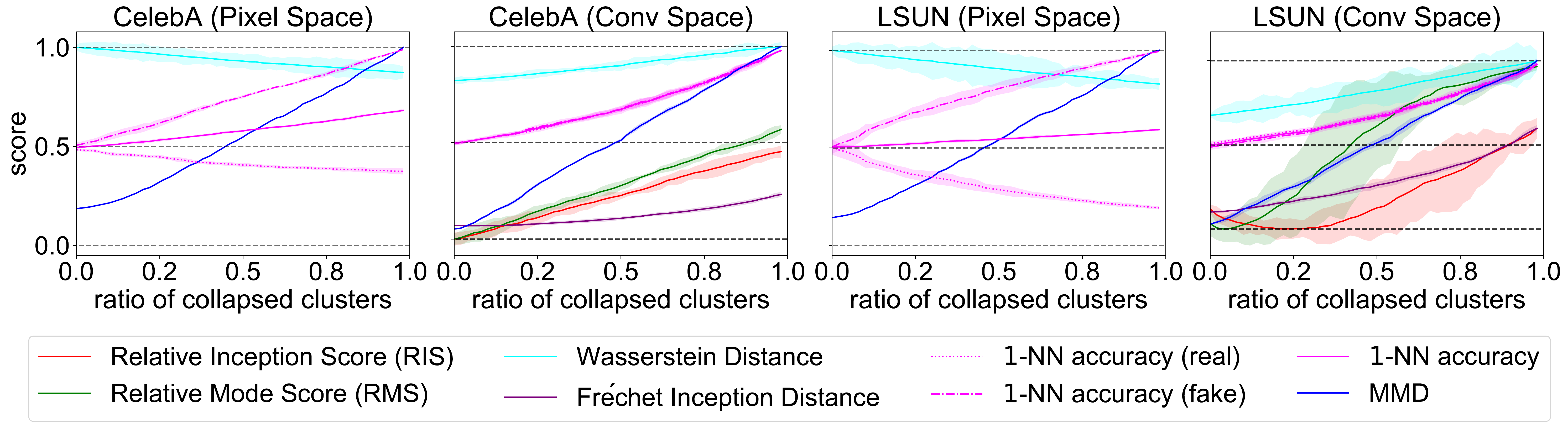}
	\vspace{-1 ex}
    \caption{Experiment on simulated mode collapsing. A metric score should increase to reflect the mismatch between true distribution and generated distribution as more modes are collapsed towards their cluster center.
		All metrics respond correctly in convolutional space. In pixel space, both Wasserstein distance and 1-NN accuracy (real) fail as they decrease in response to more collapsed clusters.}
	\label{fig:collapse}
	\vspace{-5 ex}
\end{figure*}

\paragraph{Mode collapsing and mode dropping.}
In realistic settings, $\mathbb{P}_r$ is usually very diverse since natural images are inherently multimodal. Many have conjectured that $\mathbb{P}_g$ differs from $\mathbb{P}_r$ by \emph{reducing diversity}, possibly due to the lack of model capacity or inadequate optimization \citep{arora2017generalization}. This is often manifested itself for generative models in a mix of two ways: \emph{mode dropping}, where some hard-to-represent modes of $\mathbb{P}_r$ are simply ``ignored" by $\mathbb{P}_g$; and \emph{mode collapsing}, where several modes of $\mathbb{P}_r$ are ``averaged" by $\mathbb{P}_g$ into a single mode, possibly located at a midpoint. An ideal metric should be sensitive to these two phenomena.

To test for \emph{mode collapsing}, we first randomly sample both $S_r$ and $S_r'$ as two disjoint sets of 2000 \emph{real} images. Next, we find 50 clusters in the whole training set with $k$-means and progressively replace each cluster by its respective cluster center to \emph{simulate} mode collapse. \autoref{fig:collapse} shows computed values of $\hat{\rho}(S_r, S_r')$ as the number of replaced (collapsed) clusters, denoted as $C$ increases.
%Specifically, for each increment on the $x$-axis, we randomly select one cluster (without replacement throughout the increments) and update $S_g$ by
%substituting all of its images in $S_g$ with the cluster center, then compute $\hat{\rho}(S_r, S_g)$ for all metrics on the $y$-axis.
Ideally, we expect the scores increase as $C$ grows.
We first observe that all the metrics are able to respond correctly when distances are computed in the \emph{convolutional feature space}. However, the \emph{Wasserstein} metric (cyan curve) breaks down in pixel space, as it considers a collapsed sample set (with $C\!>\!0$) being closer to the real sample set than another set of real images (with $C\!=\!0$). Moreover, although the overall 1-NN accuracy (solid magenta curve) follows the desired trend, the real and fake parts follow opposite trends: 1-NN \emph{real} accuracy (dotted magenta curve) decreases while 1-NN \emph{fake} accuracy (dashed magenta curve) increases. Again, this is inline with our explanation given in the last paragraph of Section~\ref{subsec:metrics}.
%This may also explain why in \citet{arjovsky2017wasserstein}, a GAN with MLP discriminator is much more difficult to train compared to one with a convolutional network, due to the discriminator capturing pixel-space distances.

%These results are also consistent with empirical observations, that  Wasserstein GANs can be easily fooled by mode collapsing in pixel space~\citep{arjovsky2017wasserstein}.

\begin{figure*}[t]
	\centering
    \includegraphics[width=0.9\textwidth]{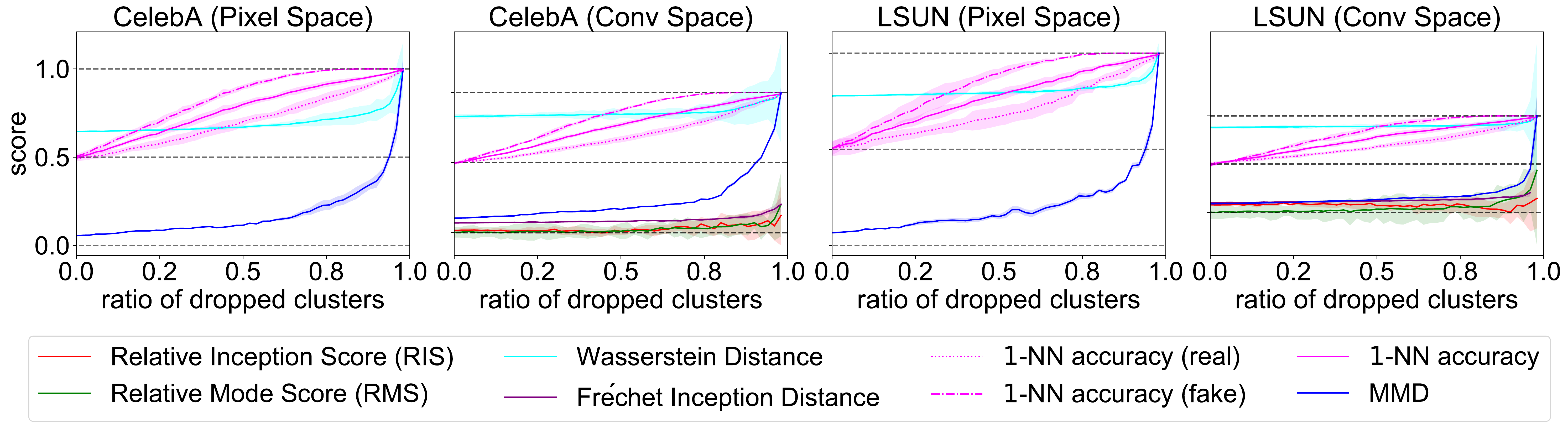}
	\vspace{-1 ex}
    \caption{Experiment on simulated mode dropping. A metric score should increase to reflect the mismatch between true distribution and generated distribution as more modes are dropped.
		All metrics except RIS and RMS respond correctly, as they only increase slightly in value even when almost all modes are dropped.}
	\label{fig:drop}
	\vspace{-2 ex}
\end{figure*}

%\paragraph{Mode dropping.}
To test for \emph{mode dropping}, we take $S_r$ as above and construct $S_r'$ by randomly removing clusters.
To keep the size of $S_r'$ constant, we replace images from the removed cluster with images randomly selected from the remaining clusters.
\autoref{fig:drop} shows how different metrics react to the number of removed clusters, also denoted as $C$.
All scores effectively discriminate against mode dropping \emph{except the RIS and RMS} - they remain almost indifferent when some modes are dropped. Again, this is perhaps caused by the fact that the Inception/Mode Score were originally designed for datasets with classes overlapping with the ImageNet dataset, and they do not generalize well to other datasets.

\begin{figure*}[!t]
	\centering
	\includegraphics[width=0.9\textwidth]{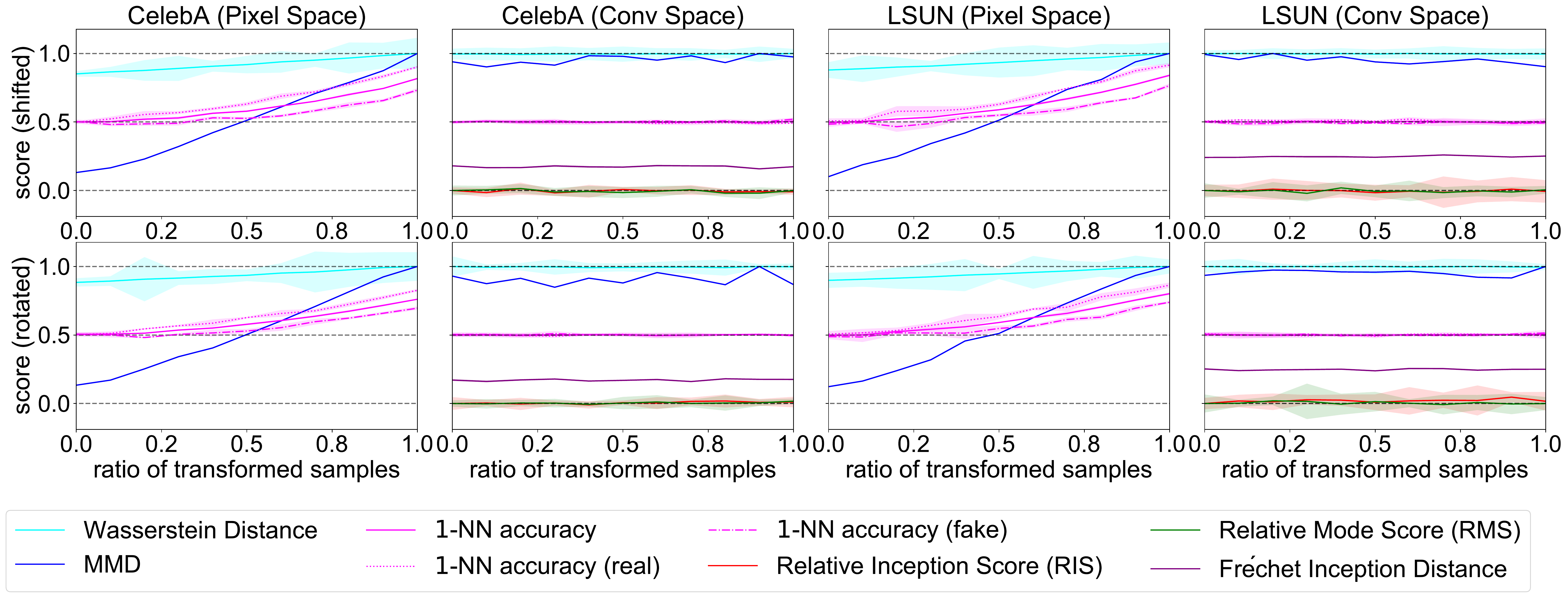}
	\vspace{-1 ex}
	\caption{
		Experiment on robustness of each metric to small transformations (rotations and translations). All metrics should remain constant across all mixes of real and transformed real samples, since the transformations do not alter image semantics.
		All metrics respond correctly in convolutional space, but not in pixel space. This experiment illustrates the unsuitability of distances in pixel space.
	}
	\label{fig:shift}
	\vspace{-4 ex}
\end{figure*}

\begin{figure*}[t]
	\centering
	\includegraphics[width=0.9\textwidth]{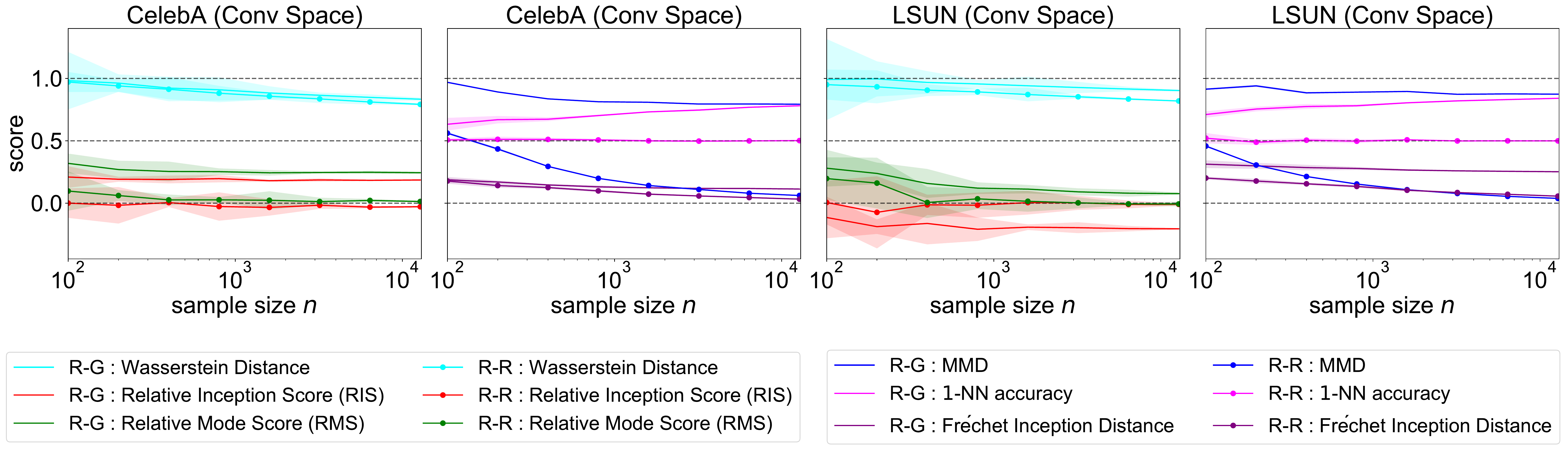}
	\vspace{-1 ex}
	\caption{The score of various metrics as a function of the number of samples. An ideal metric should result in a large gap between the real-real (R-R; $\hat{\rho}(S_r, S_r')$) and real-fake (R-G; $\hat{\rho}(S_r, S_g)$) curves in order to distinguish between real and fake distributions using as few samples as possible.
		Compared with Wasserstein distance, MMD and 1-NN accuracy require much fewer samples to discriminate real and generated images, while RIS totally fails on LSUN as it scores generated images even better (lower) than real images.}
	\label{fig:sample-size}
	\vspace{-2 ex}
\end{figure*}

\vspace{-1 ex}
\subsection{Robustness to transformations}
\vspace{-1 ex}
GANs are widely used for image datasets, which have the property that certain transformations to the input do not change its semantic meaning. Thus an ideal evaluation metric should be invariant to such transformations to some extent. For example, a generator trained on CelebA should not be penalized by a metric if its generated faces are shifted by a few pixels or rotated by a small angle.

\autoref{fig:shift} shows how the various metrics react to such small transformation to the images. In this experiment, $S_r$ and $S_r'$ are two disjoint sets of 2000 \emph{real} images sampled from the training data. However, a proportion of images from $S_r'$ are randomly shifted (up to 4 pixels) or rotated (up to 15 degrees). We can observe from the results that metrics operating in the convolutional space (or softmax space for RIS and RMS) are robust to these transformations, as all the scores are almost constant as the ratio of transformed samples increases. This is not that surprising as convolutional networks are well know for being invariant to certain transformations \citep{mallat2016understanding}.
In comparison, in the pixel space all the metrics consider the shifted/rotated images as drawn from a different distribution, highlighting the importance of computing distances in a proper feature space.

\vspace{-2 ex}
\subsection{Efficiency}
\vspace{-1 ex}
A practical GAN evaluation metric should be able to compute ``accurate'' scores from a \emph{reasonable number} of samples and within an affordable computation cost, such that it can be computed, for example, after each training epoch to monitor the training process.

\paragraph{Sample efficiency.} Here we measure the sample efficiency of various metrics by investigating how many samples are needed for each of them in order to discriminate a set of generated samples $S_g$ (from DCGAN) from a set of real samples $S_r'$. To do this, we introduce a reference set $S_r$, which is also uniformly sampled from the real training data, but is disjoint with $S_r'$. All three sample sets have the same size, i.e., $|S_r|=|S_r'|=|S_g|=n$.
We expect that an ideal metric $\rho$ should correctly score $\hat{\rho}(S_r, S_r')$ lower than $\hat{\rho}(S_r, S_g)$ with a relatively small $n$.
In other words, the number of samples $n$ needed for the metric to distinguish $S_r'$ and $S_g$ can be viewed as its sample complexity.

\begin{wrapfigure}{r}{0.3\textwidth}
	\vspace{-2.5 ex}
	\centering \includegraphics[width=0.3\textwidth]{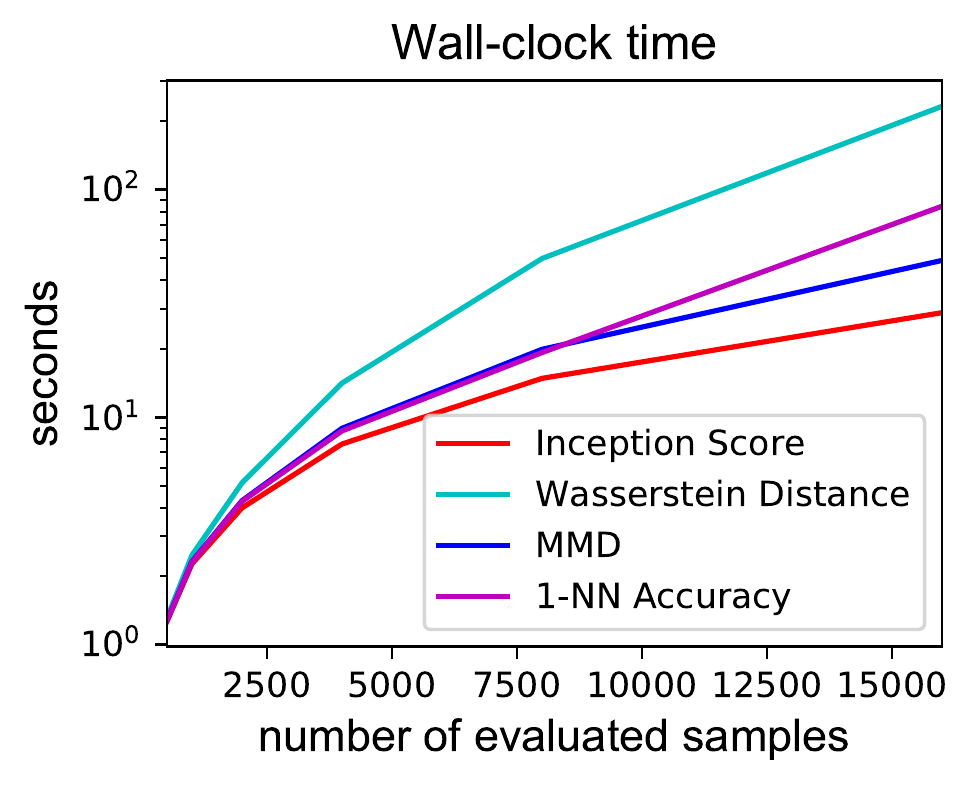}\\
	\vspace{-2 ex}
		\caption{Measurement of wall-clock time for computing various metrics as a function of the number of samples. All metrics are practical to compute for a sample of size 2000, but Wasserstein distance does not scale to large sample sizes.}
	\vspace{-1 ex}
	\label{fig:wall-time}
\end{wrapfigure}

In \autoref{fig:sample-size} we show the individual scores as a function of $n$. We can observe that MMD, FID and 1-NN accuracy computed in convolution feature space are able to distinguish the two set of images $S_g$ (solid blue and magenta curves) and $S_r'$ (dotted solid blue and magenta curves) with relatively few samples. The Wasserstein distance (cyan curves) is not discriminative with samples size less than 1000, while the RIS even considers the generated samples to be more ``real'' than the real samples on the LSUN dataset (the red curves in the third panel).
The dotted lines in \autoref{fig:sample-size} also quantify how fast the scores converge to their expectations as we increase the sample size.
%One can observe that the empirical estimations of 1-NN accuracy and the RIS are unbiased, while that of MMD and Wasserstein distance are biased. {\red Why is MMD biased? It seems to be converging to 0.}
Note that MMD for $\hat\rho(S_r, S_r')$ converges very quickly to zero and gives discriminative scores with few samples, making it a practical metric for comparing GAN models.

\paragraph{Computational efficiency.}
Fast computation of the metric is of practical concern as it helps researchers monitor the training process and diagnose problems early on, or perform early stopping. In \autoref{fig:wall-time} we investigate the computational efficiency of the above metrics by showing the wall-clock time (in log scale) to compute them as a function of the number of samples. For a typical number of 2000 samples, it only takes about 8 seconds to compute each of these metrics on an NVIDIA TitanX. In fact, the majority of time is spent on extracting features from the ResNet model. Only the Wasserstein distance becomes prohibitively slow for large sample sizes.

\begin{figure*}
	\centering
	\includegraphics[width=0.95\textwidth]{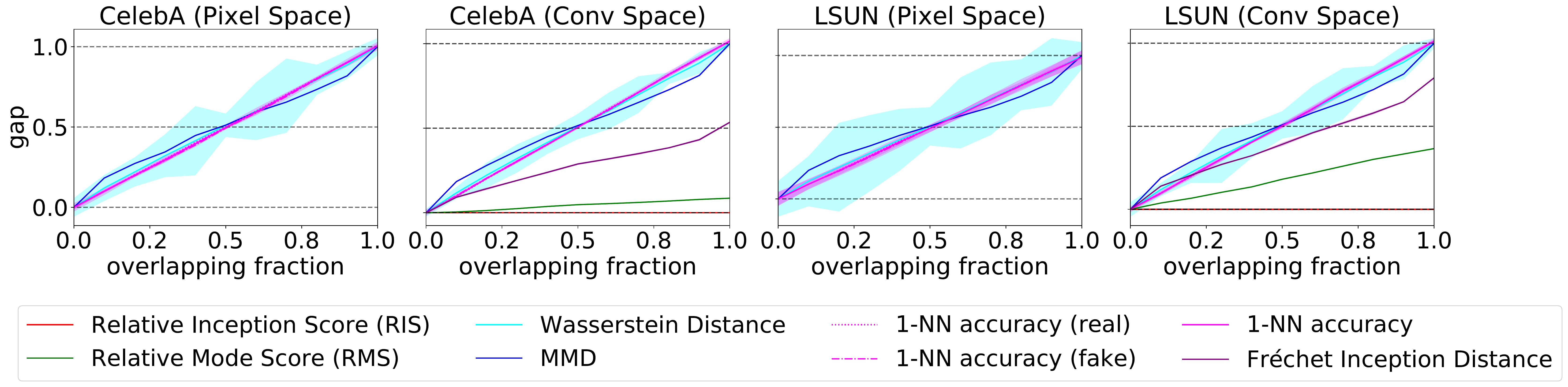}\\
\vspace{-1 ex}
%	\caption{All the metrics \emph{except RIS and RMS} are able to reveal a ``generalization gap" that detects overfitting.}
\caption{Experiment on detecting overfitting of generated samples. As more generated samples overlap with real samples from the training set, the gap between validation and training score should increase to signal overfitting.
All metrics behave correctly except for RIS and RMS, as these two metrics do not increase when the fraction of overlapping samples increases.}
	\label{fig:overfit_conv}
\vspace{-4 ex}
\end{figure*}

\vspace{-1 ex}
\subsection{Detecting overfitting}
\vspace{-1 ex}

%\begin{wrapfigure}{r}{0.3\textwidth}
%	\centering
%	\vspace{-3 ex}
%	\includegraphics[width=0.3\textwidth]{./figures/overfitting.pdf}\\
%	\vspace{-1 ex}
%	\caption{Experiment on detecting overfitting of generated samples. As more generated samples overlap with real samples from the training set, the gap between validation and training score should increase to signal overfitting.
 %          All metrics behave correctly except for RIS and RMS, as these two metrics do not increase when the fraction of overlapping samples increases.}
%	\vspace{-4 ex}
%	\label{fig:overfit_conv}
%\end{wrapfigure}

Overfitting is an artifact of training with finite samples. If a GAN successfully memorizes the training images, i.e., $\mathbb{P}_g$ is a uniform distribution over the training sample set $S_r^{tr}$, then the generated samples $S_g$ becomes a uniformly drawn set of $n$ samples from $S_r^{tr}$, and any reasonable $\hat{\rho}$ should be close to 0.
%The 1-NN two sample test has the especially appealing property that if the generator memorizes a part of $S_r$, the accuracy drops to below 0.5, which can then be interpreted as a clear signal of overfitting.
The Wasserstein distance, MMD and 1-NN two sample test are able to detect overfitting in the following sense: if we \emph{hold out} a validation set $S_r^{val}$, then $\hat{\rho}(S_g, S_r^{val})$ should be significantly higher than $\hat{\rho}(S_g, S_r^{tr})$ when $\mathbb{P}_g$ memorizes a part of $S_r^{tr}$. The difference between them can informally be viewed as a form of ``generalization gap".

We simulate the overfitting process by defining $S_r'$ as a mix of samples from the training set $S_r^{tr}$ and a second holdout set, disjoint from both $S_r^{tr}$ and $S_r^{val}$. \autoref{fig:overfit_conv} shows the gap $\hat{\rho}(S_g, S_r^{val}) - \hat{\rho}(S_g, S_r^{tr})$ of the various metrics as a function of the overlapping ratio between $S_r'$ and $S_r^{tr}$. The left most point of each curve can be viewed as the score $\hat{\rho}(S_r', S_r^{val})$ computed on a validation set since the overlap ratio is 0. For better visualization, we normalize the Wasserstein distance and MMD by dividing their corresponding score when $S_r'$ and $S_r$ have no overlap. As shown in \autoref{fig:overfit_conv}, all the metrics except RIS and RMS reflect that the ``generalization gap" increases as $S_r'$ overfits more to $S_r$. The failure of RIS is not surprising: it totally ignores the real data distribution as we discussed in Section~\ref{subsec:metrics}. While the reason that RMS also fails to detect overfitting may again be its lack of generalization to datasets with classes not contained in the ImageNet dataset. In addition, RMS operates in the softmax space, the features in which might be too specific compared to the features in the convolutional space.

%The FID also fails to detect overfitting on both datasets. We speculate that it is partly caused by the fact that FID only considers the first two order moments of the distributions. However, a thorough understanding of this phenomenon may require further investigation.

%!TEX root=main.tex

\vspace{-2 ex}
\section{Discussions and Conclusion}
\label{sec:discussion}
\vspace{-1 ex}
Based on the above analysis, we can summarize the advantages and inherent limitations of the six evaluation metrics, and conditions under which they produce meaningful results. With some of the metrics, we are able to study the problem of overfitting (see Appendix~\ref{sec:overfitting}), perform model selection on GAN models and compare them without resorting to human evaluation based on cherry-picked samples  (see Appendix~\ref{sec:compare-gan}).

\paragraph{The Inception Score} does show a reasonable correlation with the quality and diversity of generated images, which explains the wide usage in practice. However, it is ill-posed mostly because it only evaluates $\mathbb{P}_g$ as an image generation model rather than its similarity to $\mathbb{P}_r$.
Blunt violations like mixing in natural images from an entirely different distribution completely deceives the Inception Score. As a result, it may encourage the models to simply learn
sharp and diversified images (or even some adversarial noise), instead of  $\mathbb{P}_r$. This also applies to the \textbf{Mode Score}.
Moreover, the Inception Score is unable to detect overfitting since it cannot make use of a holdout validation set.

\textbf{Kernel MMD} works surprising well when it operates in the feature space of a pre-trained ResNet. It is always able to identify generative/noise images from real images, and both its sample complexity and computational complexity are low. Given these advantages, even though MMD is biased, we recommend its use in practice.

\paragraph{Wasserstein distance} works well when the distance is computed in a suitable feature space. However, it has a high sample complexity, a fact that has also been observed by \citep{arora2017generalization}. Another key weakness is that computing the exact Wasserstein distance has a time complexity of $O(n^3)$, which is prohibitively expensive as sample size increases. Compared to other methods, Wasserstein distance is less appealing as a \emph{practical evaluation metric}.

\paragraph{Fr\'echet Inception Distance} performs well in terms of discriminability, robustness and efficiency. It serves as a good metric for GANs, despite only modeling the first two moments of the distributions in feature space.

%Although fast approximation methods exist \citep{aude2016stochastic,cuturi2014fast}, it's difficult to estimate how many more samples we need to compensate for the approximation error introduced.

%The only limitation is that even when we learn the perfect distribution, the distance is not zero when we only have finitely many samples.

\textbf{1-NN classifier} appears to be an ideal metric for evaluating GANs. Not only does it enjoy all the advantages of the other metrics, it also outputs a score in the interval $[0,1]$, similar to the accuracy/error in classification problems. When the generative distribution perfectly match the true distribution, perfect score (i.e., $50\%$ accuracy) is attainable.
From \autoref{fig:mix-ratio}, we find that typical GAN models tend to achieve lower LOO accuracy for \emph{real} samples (\emph{1-NN accuracy (real)}), while higher LOO accuracy for \emph{generated} samples  (\emph{1-NN accuracy (fake)}).
This suggests
that GANs are able to capture \emph{modes} from the training distribution, such that the majority of training samples distributed around the mode centers have their nearest neighbor from the generated images, yet most of the generated images are still surrounded by generated images as they are collapsed together.
The observation indicates that the \emph{mode collapse} problem is prevalent for typical GAN models. We also note that this problem, however, cannot be effectively detected by human evaluation or the widely used Inception Score.

Overall, our empirical study suggests that the choice of feature space in which to compute various metrics is crucial. In the convolutional space of a ResNet pretrained on ImageNet, both MMD and 1-NN accuracy appear to be good metrics in terms of discriminability, robustness and efficiency. Wasserstein distance has very poor sample efficiency, while Inception Score and Mode Score appear to be unsuitable for datasets that are very different from ImageNet. We will release our source code for all these metrics, providing researchers with an off-the-shelf tool to compare and improve GAN algorithms.

Based on the two most prominent metrics, MMD and 1-NN accuracy, we study the overfitting problem of DCGAN and WGAN (in Appendix~\ref{sec:overfitting}). Despite the widespread belief that GANs are overfitting to the training data, we find that this does not occur unless there are very few training samples. This raises an interesting question regarding the generalization of GANs in comparison to the supervised setting. We hope that future work can contribute to explaining this phenomenon.

%\begin{figure}[!t]
%	\centering
%	\includegraphics[width=0.65 \columnwidth]{figures/cifar10.pdf}\\
%	\caption{Using feature spaces from a ResNet trained on CIFAR-10 to evaluate the a GAN model trained on the same dataset.}
%	\label{fig:feature-space-cifar10}
%	\vspace{-4ex}
%\end{figure}

%\begin{wrapfigure}{r}{0.3\textwidth}
%	\vspace{-1 ex}
%	\centering
%	\includegraphics[width=0.3\textwidth]{./figures/wall_time.pdf}\\
%	\vspace{-1 ex}
%		\caption{The wall-clock time to compute various metrics as a function of the number of samples.}
%	\vspace{-3 ex}
%	\label{fig:wall-time}
%\end{wrapfigure}

%\input{conclusion}

\paragraph{Acknowledgements.}
The authors are supported in part by grants from the National Science Foundation (III-1525919, IIS-1550179, IIS-1618134, S\&AS 1724282, and CCF-1740822), the Office of Naval Research DOD (N00014-17-1-2175), and the Bill and Melinda Gates Foundation. We are thankful for generous support by SAP America Inc.

\bibliography{citations}
\bibliographystyle{iclr2018_conference}

\newpage
\appendix

\section{GAN Variants used in our experiments}
\vspace{-1ex}
\label{sec:gan-variants}
Many GAN variants have been proposed recently. In this paper we consider several of them, which we briefly review in this section.

\paragraph{DCGAN}~\citep{radford2015unsupervised}.
The generator of a DCGAN takes a
lower dimensional input from a uniform noise distribution,
then projects and reshapes it to a small convolutional representation with many feature maps. After applying a series of four fractionally-strided convolutions, the generator converts this representation into a $64 \times 64$ pixel image. DCGAN is optimized by minimizing the Jensen-Shannon divergence between the real and generated images.

\paragraph{WGAN}~\citep{arjovsky2017wasserstein}. A critic network that outputs unconstrained real values is used in place of the discriminator. When the critic is Lipschitz,
this network approximates the Wasserstein distance between $S_r$ and $S_g$. A Lipschitz condition is enforced by clipping the critic networks' parameters to stay within a predefined bounding box.

\paragraph{WGAN with gradient penalty}~\citep{gulrajani2017improved} improves upon WGAN by enforcing the Lipschitz condition with a gradient penalty term.
This method significantly improves the convergence speed and the quality of the images generated by a WGAN.

\paragraph{LSGAN}~\citep{mao2016least}. Least Squares GAN adopts the least squares loss function instead of the commonly used sigmoid cross entropy loss for the discriminator,  essentially minimizing the Pearson $\chi^2$ divergence between the real distribution $\mathbb P_r$ and generative distribution $\mathbb P_g$.

\begin{figure*}[!h]
	\centering
	\includegraphics[width=0.95 \textwidth]{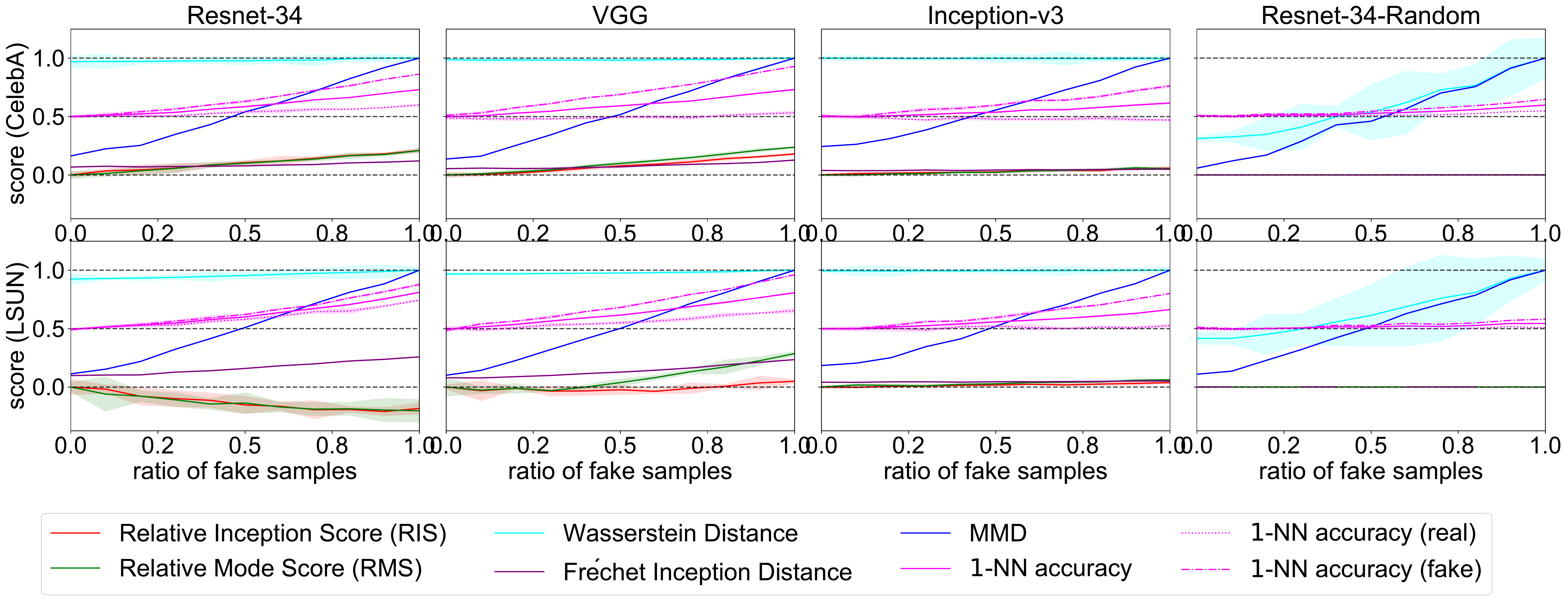}\\
	\caption{Comparison of all metrics in different feature spaces. When using different trained networks, the trends of all metrics are very similar.
	Most metrics work well even in a random network, but Wasserstein distance has very high variance and the magnitude of increase for 1-NN accuracy is small.}
	\label{fig:feature-space}
%	\vspace{-4ex}
\end{figure*}

\vspace{-1ex}
\section{The choice of feature space}
\vspace{-1ex}

The choice of features space is crucial for all these metrics. Here we consider several alternatives to the convolutional features from the 34-layer ResNet trained on ImageNet. In particular, we compute various metrics using the features extracted by (1) the VGG and Inception networks; (2) a 34-layer ResNet with random weights; (3) a ResNet classifier trained on the same dataset as the GAN models. We use the features extracted from these models to test all metrics in the discriminative experiments we performed in Section \ref{experiments}. All experimental settings are identical except for the third experiments, which is performed on the CIFAR-10 dataset \citep{cifar} instead as we need class labels to train the classifier. Note that we consider setting (3) only for analytical purposes. It is not a practical choice as GANs are mainly designed for unsupervised learning and we should not assume the existence of ground truth labels.

\begin{figure*}%{r}{0.8\textwidth}
	\centering
	\includegraphics[width=0.8 \textwidth]{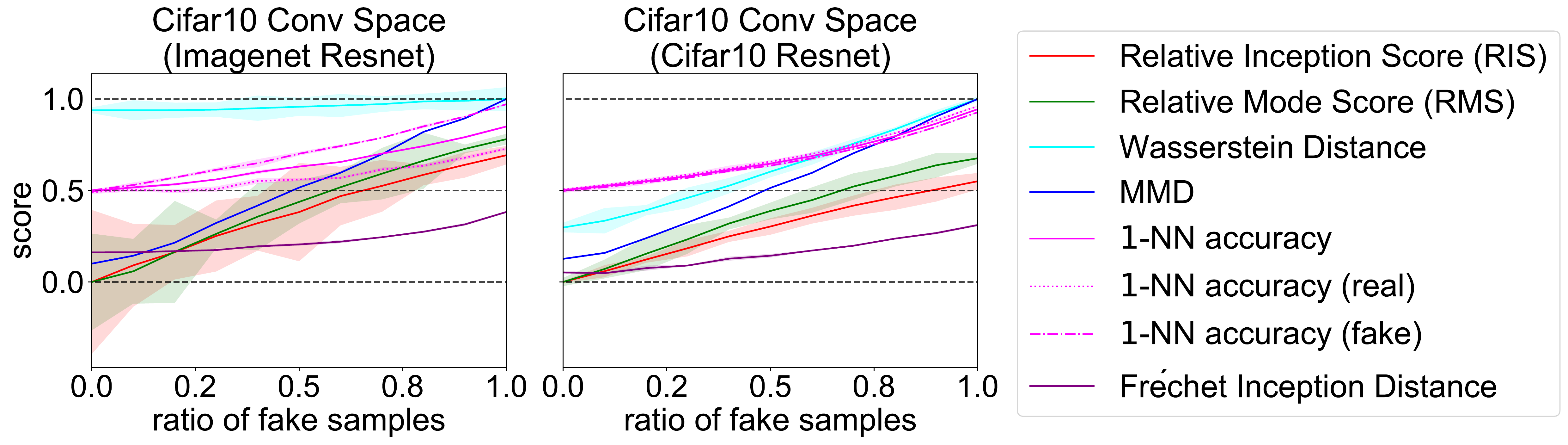}\\
	\caption{Using features extracted from a ResNet trained on CIFAR-10 (right plot) to evaluate a GAN model trained on the same dataset.
	Compared to using an extractor trained on ImageNet, the metrics appear to have lower variance. However, this may due to the feature dimensionality being smaller for CIFAR-10.}
	\label{fig:feature-space-cifar10}
\end{figure*}

The results are shown in \autoref{fig:feature-space} and \autoref{fig:feature-space-cifar10}, from which several observations can be made: (1) switching from ResNet-34 to VGG or Inception has little effect to the metric scores; (2) the features from a random network still works for MMD, while it makes the Wasserstein distance unstable and 1-NN accuracy less discriminative. Not surprisingly, the Inception Score and Mode Score becomes meaningless if we use the softmax values from the random network; (3) features extracted from the classifier trained on the same dataset as the GAN model also offers high discriminability for these metrics, especially for the Wasserstein distance. However, this may be simply due to the fact that the feature dimensionality of the ResNet trained on CIFAR-10 is much smaller than that of the ResNet-34 trained on ImageNet (64 v.s. 512).

\begin{figure*}
	\centering
	\includegraphics[width=0.95 \textwidth]{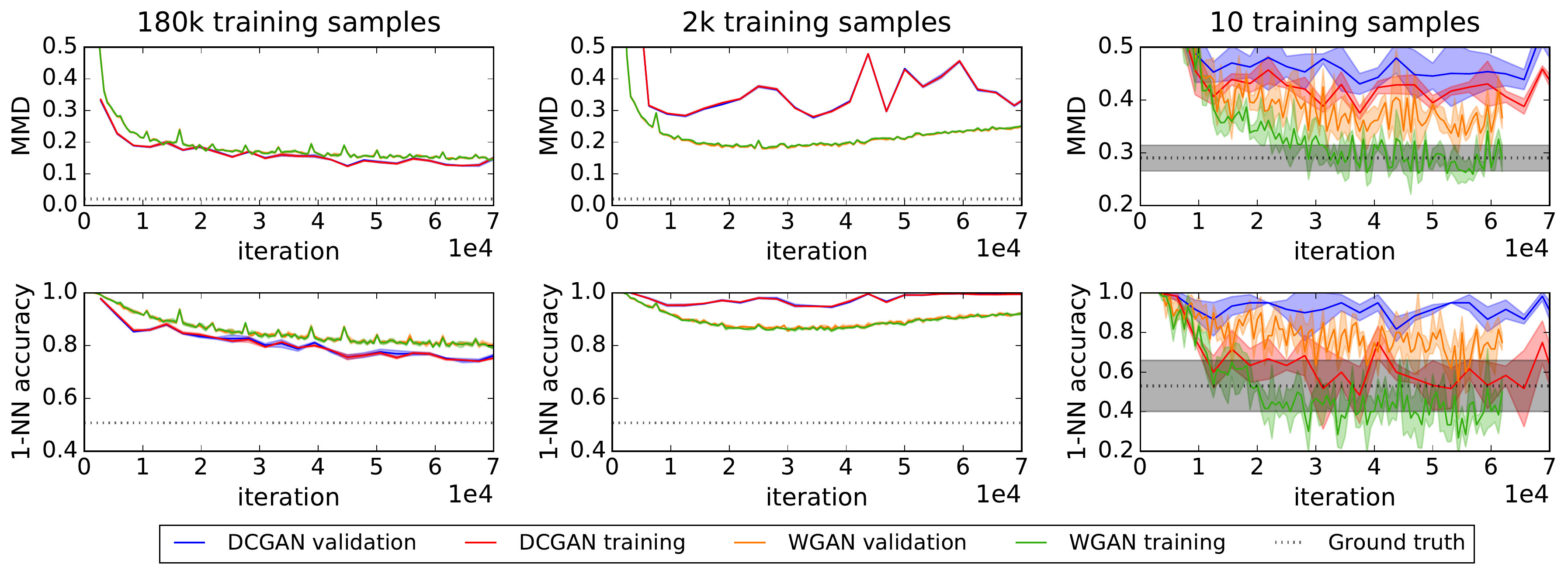}\\
	\caption{Training curves of DCGAN and WGAN on a large (\emph{left} two panels), small (\emph{middle} two panels) and tiny (\emph{right} two panels) subsets of CelebA. Note that for the first four plots, blue (yellow) curves almost overlap with the red (green) curves, indicating no overfitting detected by the two metrics. Overfitting only observed on the tiny training set, with MMD score and 1-NN accuracy significantly worse (higher) on the validation set.}
	\label{fig:tr-val-compare}
%	\vspace{-4ex}
\end{figure*}

\vspace{-1ex}
\section{Are GANs overfitting to the training data?}
\vspace{-1ex}
\label{sec:overfitting}

We trained two representative GAN models, DCGAN
%\footnote{https://github.com/pytorch/examples/tree/master/dcgan}
\citep{radford2015unsupervised} and WGAN
%\footnote{https://github.com/martinarjovsky/WassersteinGAN}
\citep{arjovsky2017wasserstein} on the CelebA dataset. Out of the $\sim$200,000 images in total, we holdout 20,000 images for validation, and the rest for training. As the training set is sufficiently large, which makes overfitting unlikely to occur, we also create a small training set and a tiny training set respectively with only 2000 and 10 images sampled from the full training set.

The training setting for DCGAN and WGAN strictly follow their original implementation, except that we change the default number of training iterations such that both models are sufficiently updated.
%We compute MMD and 1-NN accuracy based on two sets of samples drawn respectively from the real distribution and the generative distribution.
%\begin{wrapfigure}{r}{0.45\textwidth}
%	\centering
%	\vspace{-2 ex}
%	\includegraphics[width=0.45\textwidth]{figures/memorizing10.pdf}\\
%	%\includegraphics[width=0.5\textwidth]{figures/visual_earlyexit_cifar10-2.pdf}
%		\vspace{-1.5 ex}
%	\caption{DCGAN memorizing 10 images. $Top$ row: real images; $Bottom$ row: generated images.}
%	\vspace{-2 ex}
%	\label{fig:memorize}
%\end{wrapfigure}
For each metric, we compute their score on 2000 real samples and 2000 generated samples, where the real samples are drawn from either the training set or the validation set, giving rise to training and validation scores.
The results are shown in \figurename~\ref{fig:tr-val-compare}, from which we can make several observations:

\begin{itemize}[leftmargin=3ex,noitemsep,nolistsep]
	\item The training and validation scores almost overlap with each other with 2000 or 180k training samples, showing that both DCGAN and WGAN do not overfit to the training data under of these metrics. Even when using only 2000 training samples, there is still no significant difference between the training score and validation score. This shows that the training process of GANs behaves quite differently from those of supervised deep learning models, where a model can easily achieve 0 training error while behaving like random guess on the validation set \citep{zhang2016understanding}.
%In \autoref{fig:memorize}, we show that only in the extreme case with 10 training samples, overfitting occurs.
\footnote{We observed that even memorizing $50$ images is difficult for GAN models.}

	\item DCGAN outperforms WGAN on the full training set under both metrics, and converges faster. However, WGAN is much more stable on the small training set, and converges to better positions.
\end{itemize}

\begin{table*}[]
	\centering
	\small
	\caption{Comparison of several GAN models on the LSUN dataset}
	\label{table:GANLsunResults}
	\begin{tabular}{ll|r|rrrr}
		&                     & \multicolumn{1}{l|}{Real} & \multicolumn{1}{l}{DCGAN} & \multicolumn{1}{l}{WGAN} & \multicolumn{1}{l}{WGAN-GP} & \multicolumn{1}{l}{LSGAN} \\ \hline
%		Pixel Space & MMD                 & 0.0186                    & 0.0616                    & 0.0271                   & \textbf{0.0260}             & 0.0425                    \\
%		& kNN Accuracy        & 0.4969                    & 0.5735                    & 0.5685                   & \textbf{0.5070}             & 0.5305                    \\
%		& kNN Accuracy (real) & 0.4837                    & 0.4285                    & \textbf{0.3025}          & 0.5270                      & 0.4745                    \\
%		& kNN Accuracy (fake) & 0.5100                    & 0.7185                    & 0.8345                   & \textbf{0.4870}             & 0.5865                    \\ \hline
%		Conv Space  & MMD                 & 0.0188                    & 0.2046                    & 0.2703                   & \textbf{0.1935}             & 0.2316                    \\
%		& kNN Accuracy        & 0.4990                    & 0.8250                    & 0.9203                   & \textbf{0.8120}             & 0.8708                    \\
%		& kNN Accuracy (real) & 0.4949                    & \textbf{0.7585}           & 0.8800                   & 0.7645                      & 0.8035                    \\
%		& kNN Accuracy (fake) & 0.5031                    & 0.8915                    & 0.9605                   & \textbf{0.8595}             & 0.9380
		Conv Space  & MMD                 & 0.019                    & 0.205                    & 0.270                   & \textbf{0.194}             & 0.232                    \\
		& 1-NN Accuracy        & 0.499                    & 0.825                    & 0.920                   & \textbf{0.812}             & 0.871                    \\
		& 1-NN Accuracy (real) & 0.495                    & \textbf{0.759}           & 0.880                   & 0.765                      & 0.804                    \\
		& 1-NN Accuracy (fake) & 0.503                    & 0.892                    & 0.961                   & \textbf{0.860}             & 0.938 \\\hline
	\end{tabular}
	\vspace*{-2ex}
\end{table*}

\section{Comparison of popular GAN models based on quantitative evaluation metrics}
\label{sec:compare-gan}
\vspace{-1ex}
Based on our analysis, we chose MMD and 1-NN accuracy in the feature space of a 34-layer ResNet trained on ImageNet to compare several state-of-the-art GAN models. All scores are computed using 2000 samples from the holdout set and 2000 generated samples.
The GAN models evaluated include DCGAN \citep{radford2015unsupervised}, WGAN \citep{arjovsky2017wasserstein}, WGAN with gradient penalty (WGAN-GP ) \citep{gulrajani2017improved}, and LSGAN \citep{mao2016least} , all trained on the CelebA dataset. The results are reported in \autoref{table:GANLsunResults}, from which we highlight three observations:
\begin{itemize}
  \item WGAN-GP performs the best under most of the metrics.
  \item DCGAN achieves 0.759 overall 1-NN accuracy on real samples, slightly better than 0.765 achieved by WGAN-GP; while the 1-NN accuracy on generated (fake) samples achieved by DCGAN is higher than that by WGAN-GP (0.892 \emph{v.s.} 0.860).  This seems to suggest that DCGAN is better at capturing modes in the training data distribution, while its generated samples are more collapsed compared to WGAN-GP. Such subtle difference is unlikely to be discovered by the Inception Score or human evaluation.
  \item The 1-NN accuracy for all evaluated GAN models are higher than $0.8$ ,   far above the ground truth of $0.5$. The MMD score of the four GAN models are also much larger than that of ground truth ($0.019$).  This indicates that even state-of-the-art GAN models are far from learning the true distribution.
  \end{itemize}

\end{document}